\documentclass[runningheads]{llncs}

\usepackage{amsmath}
\usepackage{amsfonts}
\usepackage{booktabs}
\usepackage{graphicx}
\usepackage{multirow}
\usepackage{url}
\usepackage{orcidlink}

\makeatletter
\@ifundefined{orcidID}
  {\newcommand{\orcidID}[1]{\,\orcidlink{#1}}}
  {\renewcommand{\orcidID}[1]{\,\orcidlink{#1}}}
\makeatother
\usepackage{microtype}
\usepackage{float}
\usepackage{placeins}
\usepackage{etoolbox}
\AtBeginEnvironment{thebibliography}{\footnotesize\setlength{\itemsep}{-3pt}\setlength{\parsep}{0pt}\setlength{\parskip}{0pt}}
\makeatletter
\patchcmd{\thebibliography}{\small}{\footnotesize}{}{}
\renewcommand\subsection{\@startsection{subsection}{2}{\z@}%
                       {-17\p@ \@plus -4\p@ \@minus -4\p@}%
                       {8\p@ \@plus 4\p@ \@minus 4\p@}%
                       {\normalfont\normalsize\bfseries\boldmath
                        \rightskip=\z@ \@plus 8em\pretolerance=10000 }}
\makeatother
\usepackage{marvosym}
\newcommand{\method}{RSC-GestureNet}
\newcommand{\base}{RSC-Gesture-Pose}
\newcommand{\datasetc}{CTPGesture-C}
\newcommand{\arvariant}{RSC-GestureNet-AR}

\newcommand{\metricup}{\ensuremath{\uparrow}}
\newcommand{\metricdown}{\ensuremath{\downarrow}}
\renewcommand{\arraystretch}{0.90}

\begin{document}

\title{RSC-GestureNet: Reliability-Aware Selective Causal Recognition of Chinese Traffic Police Gestures}
\titlerunning{RSC-GestureNet for Causal Traffic Police Gesture Recognition}

\author{
Cheng Li\inst{1}\textsuperscript{*\,\Letter}
\orcidID{0009-0009-4322-7591}
\and
Renjun Gao\inst{2}\textsuperscript{*\,\Letter}
\orcidID{0009-0005-0028-428X}
\and
Boyi Fu\inst{2}
\orcidID{0009-0007-6704-2175}
}

\authorrunning{C. Li, R. Gao, and B. Fu}

\institute{
Academy of Interdisciplinary Studies,
The Hong Kong University of Science and Technology,
Hong Kong SAR, China\\
\email{\href{mailto:clieo@connect.ust.hk}{clieo@connect.ust.hk}}
\and
Faculty of Innovation Engineering,
Macau University of Science and Technology,
Macao SAR, China\\
\email{\{\href{mailto:1220000217@student.must.edu.mo}{1220000217},\href{mailto:1220022600@student.must.edu.mo}{1220022600}\}@student.must.edu.mo}\\[1mm]
\textsuperscript{*} Equal Contribution \qquad \Letter\ Corresponding Author
}

\maketitle
\vspace*{-3mm}
\setcounter{footnote}{0}
\begin{abstract}
Traffic police gestures are safety-critical perception cues for autonomous driving. A deployable recognizer must infer commands causally from continuous full-frame video, remain stable around transitional arm motion, and avoid over-trusting corrupted pose measurements. This study presents \method, a reliability-aware selective causal recognizer, for Chinese traffic police gestures. The model treats pose confidence as a first-class signal: unreliable joints are down-weighted during graph reasoning, temporal evidence is aggregated causally, and calibrated predictions are selectively emitted through a reliability-aware inference rule. We further introduce \datasetc, a reproducible feature-level corruption benchmark with seven pose/RGB degradation families, and an RGB-level diagnostic in which corrupted frames are reprocessed by MediaPipe before recognition. On the complete official CTPGesture v1 split (134,424 labeled frames and 33,451 causal windows), \method\ achieves 93.33$\pm$0.24\% accuracy, 91.71$\pm$0.27\% macro-F1, 91.69$\pm$0.29\% online macro-F1, 98.80$\pm$0.07\% Early@10, 0.153$\pm$0.013 s TTC, and the best robust macro-F1 among evaluated methods. Under the same split and causal protocol, it exceeds reproduced traffic-specific MD-GCN and HLP-GCN baselines by 3.23–4.11 macro-F1 points and 2.15–3.07 online-F1 points. These results, together with calibration, selective-risk, statistical, adaptive-branching, and image-level re-extraction analyses, indicate that explicit pose-reliability modeling improves early, stable, and robust traffic-command recognition. The project is available at \url{https://github.com/chengli24/rsc-gesturenet-prcv2026} 
\end{abstract}

\section{Introduction}
Traffic police hand signals can override traffic lights and road signs in congested intersections, construction zones, emergency traffic control, and temporary road-management scenarios. Public full-frame traffic-police benchmarks such as CTPGesture make this command-perception problem concrete for vision systems \cite{ctpgesture}. For an autonomous vehicle, recognizing such commands is not a peripheral task: a missed stop command or an unstable turn command can directly affect planning. The task is also harder than trimmed gesture classification because the recognizer observes continuous road-scene video, must update predictions from past frames only, and should suppress command flicker while the officer moves through intermediate poses.

Vision-based traffic-gesture recognition has progressed from handcrafted skeleton descriptors and recurrent classifiers \cite{cpmtraffic} to graph-based models that exploit body topology and command-specific arm relations. MD-GCN uses multichannel dilated graph reasoning \cite{mdgcn}; natural-scene GCN variants adapt skeleton graphs to road backgrounds \cite{gcnnatural2021}; height-layered and position-rotation graph designs encode traffic-command priors \cite{hlpgcn,prgraph}; and recent lightweight fusion models emphasize practical spatiotemporal recognition \cite{ntpgr}. These studies show that skeleton-centric representations are effective, but they are often reported with different splits, feature definitions, or offline protocols, making deployment-oriented comparison difficult.

Three gaps remain especially important for vehicle-facing use. First, a recognizer should be evaluated causally, because future frames are unavailable at decision time and transition delay matters; causal temporal models provide a natural basis for this setting \cite{tcn,mstcn}. Second, 2D pose is an estimated measurement rather than ground truth: modern pose pipelines expose landmark confidence \cite{openpose,mediapipe,blazepose}, but low illumination, blur, distance, and occlusion can still corrupt wrists and elbows, which carry most command semantics. Third, traffic commands are risk-asymmetric, so calibrated confidence and selective prediction are useful for deferring unstable decisions instead of forcing a potentially unsafe label \cite{calibration,selectivenet}; common-corruption studies further show that clean accuracy alone can hide robustness failures \cite{imagenetc}.

This study addresses these gaps with a reliability-centered causal recognition framework. Each prediction is generated from a 32-frame causal window and evaluated as part of an online command stream. The proposed \method\ treats pose confidence as a modeling signal: unreliable joints are down-weighted, command-critical upper-body relations are strengthened, stop-to-motion confusions are penalized during training, calibrated probabilities are used for selective emission, and branch weights adapt when global or upper-limb reliability degrades. Unlike comparisons that only quote incompatible published numbers, all controlled baselines---including MD-GCN and HLP-GCN traffic-graph reimplementations\footnote{Because the original MD-GCN and HLP-GCN model code is not publicly available, we reproduce their network structures from the corresponding papers, train them under the unified protocol, and provide our reproduction code in the project page.}---use the official CTPGesture v1 split, the same preprocessing cache, the same causal windows, and the same evaluation scripts. This isolates the effect of reliability-aware modeling from protocol changes.

\noindent\textbf{Contributions.} This paper makes four contributions:
\begin{itemize}\itemsep0pt
\item A reproducible full-frame causal protocol on CTPGesture v1, with three-seed statistics, online macro-F1, Early@10, TTC, PSR, calibration-aware post-processing, and sealed test-set reporting.
\item \method, a reliability-aware selective recognizer with confidence-gated graph reasoning, upper-body dynamics, danger-aware learning, pose--RGB fusion, vector calibration, and adaptive inference.
\item \datasetc, a feature-level corruption benchmark, plus an RGB-level pose re-extraction diagnostic for low-light and motion-blur stress.
\item A controlled evaluation showing that \method\ improves over generic baselines, calibrated baselines, and traffic-specific graph reimplementations under the same official split and causal protocol.
\end{itemize}

\section{Related Work}
\subsection{Traffic-Police Gesture Recognition}
Early traffic-police gesture recognition methods typically relied on depth sensors, wearable devices, handcrafted geometric descriptors, or skeleton features followed by recurrent classifiers. These studies established that official police commands are strongly structured by upper-body posture and temporal arm motion, but their reliance on specialized sensing or hand-designed features limits robustness in uncontrolled traffic scenes. He et al. combined a convolutional pose machine, skeletal geometry, and LSTM temporal modeling for Chinese traffic-police gestures \cite{cpmtraffic}. The CTPGesture dataset later provided full-frame traffic-police videos with frame-level labels and a public train/test organization, making reproducible vision-based evaluation more feasible \cite{ctpgesture}.

Graph-based traffic-gesture models further improve command representation by encoding body topology and spatiotemporal relations. MD-GCN uses multichannel dilated graph convolutions \cite{mdgcn}; natural-scene GCN variants adapt skeleton graph reasoning to road-background videos \cite{gcnnatural2021}; height-layered and position-rotation graph designs introduce command-oriented body partitions and relation structures \cite{hlpgcn,prgraph}; and NTPGR emphasizes efficient spatiotemporal fusion for practical recognition \cite{ntpgr}. These studies motivate skeleton-centered modeling, but many reported results are obtained under different dataset versions, feature definitions, or offline protocols. This study therefore treats published systems as methodological context and evaluates all compared models under a unified official split, causal window construction, and shared evaluation pipeline.

\subsection{Skeleton Graph Modeling and Pose Reliability}
Skeleton-based action recognition represents human motion as a spatiotemporal graph in which joints are nodes and anatomical or learned relations are edges. Compact temporal models such as GRU encode motion through gated recurrent states \cite{gru}, temporal convolutional networks aggregate past evidence with efficient causal or dilated filters \cite{tcn}, and ST-GCN introduced explicit spatial-temporal graph propagation along body joints and temporal links \cite{stgcn}. Later graph models refine topology, temporal scale, and representation quality, for example through multi-scale graph reasoning and channel-wise topology refinement \cite{msg3d,ctrgcn}, while pose heatmap and information-theoretic formulations revisit the skeleton representation itself \cite{posec3d,infogcn}.

For traffic-police commands, however, the skeleton is an estimated intermediate representation rather than a clean sensor measurement. OpenPose introduced part-affinity-field-based multi-person pose estimation \cite{openpose}, while MediaPipe/BlazePose provides efficient body tracking suitable for lightweight pipelines \cite{mediapipe,blazepose}. In road-scene videos, distance, occlusion, motion blur, illumination variation, and background clutter can suppress wrists or elbows, which are precisely the landmarks carrying command semantics. Existing robust skeleton representations partially improve feature stability \cite{posec3d,infogcn}, but they do not directly specify how landmark confidence should affect a safety-relevant command output. This study therefore makes pose reliability an explicit modeling signal through confidence-gated graph aggregation, upper-body relation enhancement, and lightweight pose--RGB fusion.

\subsection{Causal and Selective Recognition}
Many action-recognition benchmarks evaluate trimmed clips or offline sequences, whereas a vehicle-facing command recognizer must update its decision from past observations only. Causal TCNs and multi-stage temporal convolutional networks show that temporal context can be aggregated efficiently for action segmentation and detection \cite{tcn,mstcn}. For traffic control, however, recognition quality cannot be judged only by final clip accuracy. A deployable output should be early enough for planning, stable across adjacent windows, and conservative near transition poses or unreliable visual evidence.

Calibration and selective prediction provide useful principles for this deployment setting. Calibration requires confidence to be informative about correctness \cite{calibration}, selective prediction studies the risk--coverage tradeoff when uncertain inputs can be deferred rather than forced into a label \cite{selectivenet}, and common-corruption benchmarks reveal failure modes hidden by clean accuracy \cite{imagenetc}. Motivated by these lines of work, this study uses training-internal calibration with a sealed official test split, defines \datasetc\ for pose/RGB corruption analysis, and emits commands only when calibrated confidence and short-term temporal consistency satisfy a preselected rule.

\section{Methodology}
\subsection{Data Preprocessing}
The proposed framework is evaluated on the complete official CTPGesture v1 split, which contains full-frame RGB videos of traffic officers performing eight Chinese traffic-command gestures and an inactive state \cite{ctpgesture}. The official videos are recorded at 1080$\times$1080 resolution and 15 fps with frame-level labels and a fixed train/test organization. This study does not redefine the test set or remove difficult sequences. All official v1 videos are processed, resulting in 134,424 labeled frames, 133,861 successful pose detections, 16,693 training windows, and 16,758 testing windows, for a total of 33,451 full-frame causal windows.

Each RGB frame is processed by MediaPipe Pose based on BlazePose in video mode, with model complexity 0, landmark smoothing, and minimum detection/tracking confidence 0.35 \cite{mediapipe,blazepose}. Frames are resized to a maximum side length of 320 pixels before pose extraction to reduce computational cost while preserving the officer's upper-body structure. The 33 MediaPipe landmarks are mapped to 17 compact body joints, including the nose, shoulders, elbows, wrists, hips, knees, ankles, ears, and mouth-side landmarks. This compact representation retains the joints most relevant to police-command semantics while avoiding facial and redundant landmarks that are unstable in distant full-frame scenes.

For joint $j$ at frame $t$, the pose feature is
\begin{equation}
 x_{t,j}=[u_{t,j},v_{t,j},c_{t,j},\Delta u_{t,j},\Delta v_{t,j}],
\end{equation}
where $(u,v)$ are image-normalized coordinates, $c$ is the detector confidence, and $\Delta$ denotes a causal finite-difference velocity. Coordinates are centered at the torso and divided by a torso-scale estimate before windowing, reducing sensitivity to camera distance, officer position, and modest viewpoint variation. Missing landmarks are retained with low confidence instead of being removed, so that the model can distinguish absent evidence from reliable joint observations. Velocity is computed only from previous frames and clipped to suppress occasional pose-tracking spikes.

A lightweight appearance descriptor is extracted from a person-centered crop determined by visible pose landmarks with a 32-pixel margin. The crop is resized to 64$\times$64 and summarized using four-bin histograms over hue, saturation, and value, producing a 12-dimensional RGB descriptor per frame. This descriptor supplies illumination and coarse appearance cues without introducing a large image backbone. The design therefore keeps the recognition head lightweight and makes robustness analysis attributable to pose reliability, temporal modeling, and low-dimensional fusion rather than to an opaque visual encoder.

\begin{figure}[t]
\centering
\includegraphics[width=.99\linewidth]{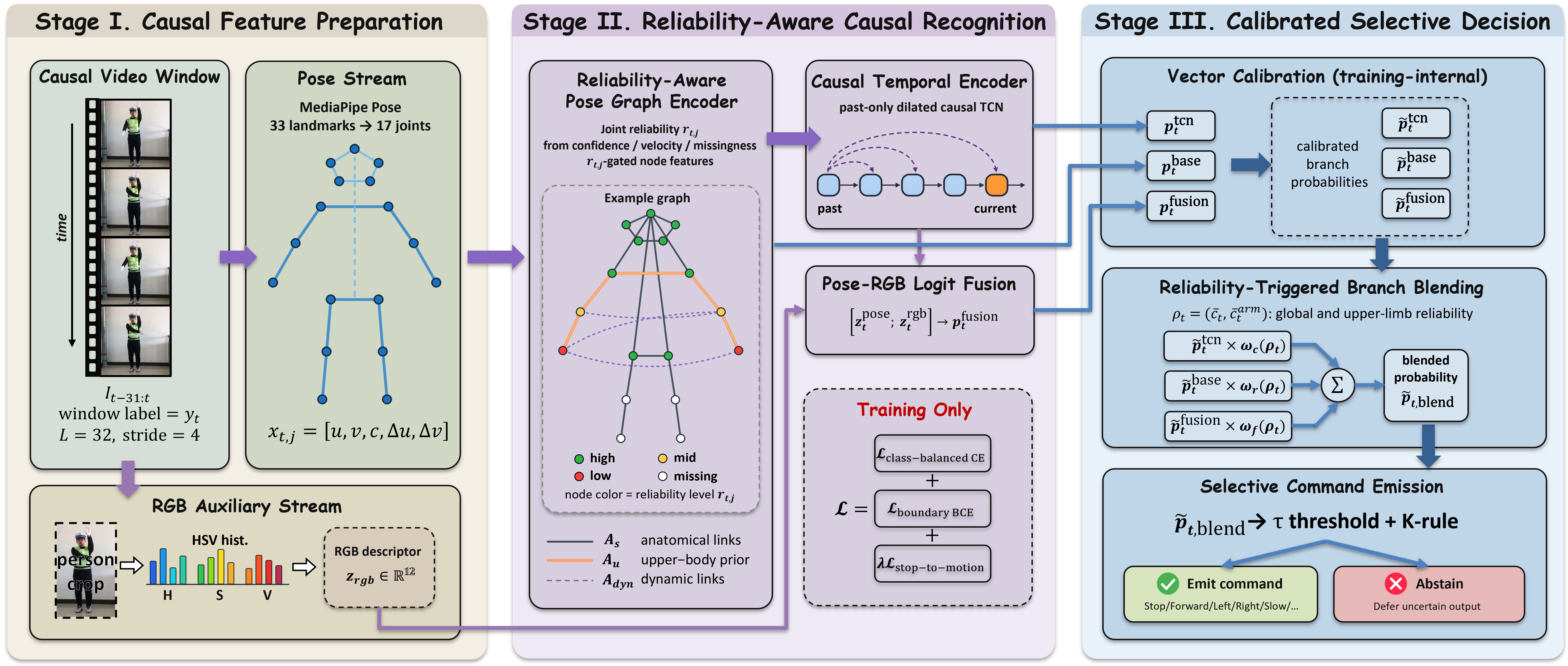}
\caption{Overview of \method. The pipeline performs reliability-aware pose preprocessing, confidence-gated graph reasoning, causal temporal modeling, danger-aware optimization, lightweight pose--RGB fusion, vector calibration, and selective command emission.}
\label{fig:method}
\end{figure}

\subsection{Causal Command Recognition Task}
Given a video stream $\{I_t\}_{t=1}^{T}$ and frame labels $y_t\in\{0,\ldots,C-1\}$, the recognizer predicts a class distribution $p_t\in\mathbb{R}^{C}$ at each causal output step. The nine classes are inactive/no-command (the official ``No gesture'' label), stop, forward, left turn, left-turn waiting, right turn, lane changing, slow down, and pull over. Recognition uses windows of length $L=32$ with stride four. Each window is labeled by its last frame, so the prediction corresponds to the command active at the current time rather than to a future segment. No future frames are used for feature construction, boundary features, temporal filtering, calibration, or stabilization.

The streaming classifier is converted into an emitted command through a confidence-and-consistency rule. A concrete command is released only when the last $K$ predictions agree and the maximum calibrated confidence exceeds a threshold; otherwise the system abstains. This abstention is an output decision and is distinct from the inactive/no-command class in the label set. Time-to-command (TTC) is measured from a non-inactive transition to the first emitted correct command and converted to seconds using the 15 fps video rate. TTC therefore measures recognition-window delay, not the full latency of video decoding, pose extraction, vehicle planning, or actuation.

\subsection{Reliability-Aware Spatial Graph Reasoning}
The spatial branch models detector reliability before graph aggregation. For each joint $j$ at time $t$, a reliability score is estimated as
\begin{equation}
 r_{t,j}=\sigma\big(f_r([c_{t,j},\lVert\Delta p_{t,j}\rVert,m_{t,j}])\big),
\end{equation}
where $m_{t,j}$ indicates missing or imputed landmarks and $f_r$ is a lightweight multilayer perceptron. The score gates node features before message passing, reducing the influence of low-confidence joints and implausible velocity spikes. This mechanism is particularly important for wrists and elbows, because small localization errors in these joints can alter the semantic distinction between stopping, turning, waiting, and lane-changing commands.

The graph adjacency combines anatomical structure, command-specific upper-body priors, and sample-dependent relations:
\begin{equation}
 A=A_s+\alpha A_u+\beta\,\operatorname{softmax}(QK^\top/\sqrt{d}),
\end{equation}
where $A_s$ is the physical skeleton adjacency, $A_u$ emphasizes shoulders, elbows, and wrists, and the dynamic term learns instance-specific coordination. Each graph block applies reliability-gated feature projection, relation-weighted aggregation, normalization, residual connection, and dropout. The static graph preserves body topology, the upper-body prior encodes traffic-command semantics, and the dynamic relation term adapts to individual execution style, partial occlusion, and viewpoint changes.

\subsection{Causal Temporal Modeling and Objective}
The temporal encoder uses dilated causal convolutions over graph features. For a hidden sequence $h_{1:t}$, the output at time $t$ depends only on $\{h_\tau\}_{\tau\le t}$, allowing the receptive field to expand without violating causality. Dilated residual blocks are suitable for traffic commands because several classes differ by short motion phases rather than a single static pose; for example, turning and waiting gestures may share similar arm configurations near transition boundaries.

The prediction head produces class logits and an auxiliary boundary logit. Boundary supervision is positive within a radius of two frames around label transitions, encouraging sensitivity to command changes while preserving causal inference. Training combines class recognition, boundary supervision, and danger-aware confusion regularization:
\begin{equation}
\mathcal{L}=\mathcal{L}_{ce}+\lambda_b\mathcal{L}_{boundary}+\lambda_d\sum_{(y,k)\in\mathcal{D}}\mathbf{1}[y_t=y]p_{t,k}.
\end{equation}
Here $\mathcal{L}_{ce}$ is class-balanced cross entropy, $\mathcal{L}_{boundary}$ is binary cross entropy for transition proximity, $\mathcal{D}$ contains operationally harmful stop-to-motion confusions such as Stop$\rightarrow$\{Forward, Left Turn, Right Turn, Lane Changing\}, and $p_{t,k}$ is the predicted probability of class $k$. This study uses $\lambda_b=0.12$ and $\lambda_d=0.18$. The regularizer does not make risky mistakes impossible; instead, it shifts probability mass away from error modes that are more consequential for downstream driving decisions.

\subsection{Pose--RGB Fusion and Reliability-Aware Inference}
The pose and RGB streams are fused at the logit level:
\begin{equation}
 p_t^{fusion}=\operatorname{softmax}(W[z_t^{pose};z_t^{rgb}]).
\end{equation}
Late fusion keeps the pose branch interpretable while allowing appearance context to help when pose confidence degrades. After training, vector calibration is fitted only on predictions from a training-internal calibration stream, and no official test labels are used to select calibration parameters, thresholds, or reliability-branch weights.

The final inference head applies reliability-triggered probability blending:
\begin{equation}
 p_t^{final}=w_f(\rho_t)p_t^{fusion}+w_r(\rho_t)p_t^{base}+w_c(\rho_t)p_t^{tcn},
\end{equation}
where $\rho_t=(\bar c_t,\bar c_t^{arm})$ summarizes global confidence and upper-limb visibility over the causal window. When reliability is high, the fusion branch dominates; when global or arm-level confidence decreases, complementary pose-only and temporal branches receive higher weight. The calibrated distribution is then passed to the stable-$K$ rule, yielding either an emitted command or a deferred output.

\section{Experiments}
\subsection{Environment Setup and Protocol}
All experiments were implemented in PyTorch on a single NVIDIA RTX A4000 GPU. Unless otherwise specified, recognizers use the official CTPGesture v1 train/test split, the same processed pose--RGB cache, 32-frame causal windows with stride four, nine classes, 17 joints, batch size 96, hidden dimension 96, dropout 0.12, AdamW with learning rate $10^{-3}$, mild label smoothing, pose augmentation, gradient clipping, a 26-epoch budget, and validation macro-F1 checkpoint selection. Results are mean $\pm$ standard deviation over three random seeds.

Macro-F1 is the primary metric because commands are imbalanced. Online-F1 is computed after stabilization; Early@10 measures whether the correct command is emitted within ten causal output steps after a transition; TTC measures delay from a non-inactive transition to the first correct command; PSR counts stabilized switches per second; and Robust-F1 averages macro-F1 over all \datasetc\ corruptions. Calibration, threshold selection, and reliability-branch weighting use only a training-internal calibration stream. The official test split is sealed and is never used for checkpoint selection, calibration, threshold tuning, or branch-weight fitting. The main online setting uses $K=1$ and threshold 0.51; a training-stream sweep shows nearby thresholds remain stable, whereas larger $K$ values reduce PSR but degrade Online-F1 and TTC.

The comparison covers class-prior, frame-level, recurrent, TCN, generic graph, calibrated, and reliability-aware recognizers. To address direct traffic-police graph baselines, we also reimplement MD-GCN \cite{mdgcn} and HLP-GCN \cite{hlpgcn,prgraph}, corresponding to multichannel dilated graph reasoning and height/position-rotation graph priors, respectively. Both use the same input representation, split, causal windows, and evaluation code, and are run for three seeds with a 26-epoch budget.

\subsection{Main Evaluation}
Table~\ref{tab:main} reports the controlled comparison under the official full-frame causal protocol. Accuracy alone is insufficient: the Majority baseline obtains 56.59\% accuracy but only 8.03\% macro-F1. Temporal and graph baselines improve recognition, yet their online-F1 remains below their clean macro-F1 because transition windows and unstable arm motion are difficult. Reliability modeling narrows this gap: online-F1 increases from 82.64\% for ST-GCN to 87.87\% for RSC and 88.50\% for \base.

The traffic-specific reimplementations make the comparison less indirect. MD-GCN and HLP-GCN baselines improve online-F1 over generic ST-GCN, confirming that command-oriented graph design is useful. However, under the same official split and causal protocol, \method\ remains clearly stronger: it exceeds MD-GCN by 3.25 accuracy points, 3.23 macro-F1 points, 2.15 online-F1 points, and 0.104 s TTC, and exceeds HLP-GCN by 4.25 accuracy points, 4.11 macro-F1 points, 3.07 online-F1 points, and 0.160 s TTC.

Calibration strengthens competitive baselines, but \method\ remains best because calibrated probabilities are coupled with reliability-triggered branch blending and selective command emission. Relative to the strongest calibrated baselines, \method\ improves accuracy by 0.24 points, macro-F1 by 0.35--0.36 points, online-F1 by 0.36 points, Early@10 by 0.39 points, TTC by 0.061 s, and Robust-F1 by 0.67 points. The clean-set margins over calibrated baselines are moderate, but the consistent gains across online behavior, command delay, calibration, selective risk, direct traffic-specific baselines, and corruption robustness support the reliability-aware design.
\begin{table}[t]
\centering
\setlength{\belowcaptionskip}{6pt}
\vspace*{1pt}
\caption{Controlled comparison on the official CTPGesture v1 split under the full-frame causal protocol. Values are mean $\pm$ standard deviation over three seeds. $\metricup$/$\metricdown$ indicate higher/lower is better. Traffic-specific reimplementations marked $\dagger$ use three seeds and a 26-epoch budget; other trainable models are also use 26 epochs. Best values are bold.}
\label{tab:main}
\scriptsize
\setlength{\tabcolsep}{2.2pt}
\resizebox{\linewidth}{!}{
\begin{tabular}{lcccccc}
\toprule
Model & Accuracy (\%)$\metricup$ & Macro-F1 (\%)$\metricup$ & Online-F1 (\%)$\metricup$ & Early@10 (\%)$\metricup$ & TTC (s)$\metricdown$ & Robust-F1 (\%)$\metricup$\\
\midrule
Majority & 56.59$\pm$0.00 & 8.03$\pm$0.00 & 8.03$\pm$0.00 & 0.00$\pm$0.00 & -- & --\\
MLP & 83.74$\pm$0.19 & 81.98$\pm$0.31 & 80.81$\pm$0.68 & 95.96$\pm$0.67 & 0.711$\pm$0.057 & --\\
GRU \cite{gru} & 88.82$\pm$0.43 & 87.10$\pm$0.45 & 81.99$\pm$0.82 & 98.14$\pm$0.31 & 0.345$\pm$0.058 & --\\
TCN \cite{tcn} & 89.88$\pm$0.54 & 88.29$\pm$0.62 & 82.32$\pm$0.16 & 97.87$\pm$0.34 & 0.377$\pm$0.043 & 87.82$\pm$0.94\\
ST-GCN \cite{stgcn} & 89.45$\pm$0.65 & 87.94$\pm$0.72 & 82.64$\pm$0.43 & 98.06$\pm$0.41 & 0.328$\pm$0.060 & 85.53$\pm$0.39\\
MD-GCN$^\dagger$ \cite{mdgcn} & 90.08$\pm$1.38 & 88.48$\pm$1.43 & 89.54$\pm$0.76 & 98.14$\pm$0.42 & 0.257$\pm$0.035 & --\\
HLP-GCN$^\dagger$ \cite{hlpgcn,prgraph} & 89.08$\pm$0.22 & 87.60$\pm$0.33 & 88.62$\pm$1.16 & 98.10$\pm$0.29 & 0.313$\pm$0.149 & --\\
RSC & 89.96$\pm$0.64 & 88.48$\pm$0.61 & 87.87$\pm$0.63 & 98.33$\pm$0.18 & 0.257$\pm$0.056 & --\\
\base & 90.19$\pm$0.29 & 88.78$\pm$0.40 & 88.50$\pm$0.51 & 98.29$\pm$0.36 & 0.284$\pm$0.133 & 87.75$\pm$0.78\\
RSC-GestureNet-Fusion & 91.38$\pm$0.38 & 89.95$\pm$0.42 & 89.28$\pm$0.53 & 98.72$\pm$0.20 & 0.191$\pm$0.040 & 88.93$\pm$0.77\\
\midrule
TCN+VC \cite{tcn} & 92.97$\pm$0.18 & 91.16$\pm$0.21 & 91.08$\pm$0.23 & 98.41$\pm$0.07 & 0.214$\pm$0.007 & 90.29$\pm$0.34\\
ST-GCN+VC \cite{stgcn} & 92.47$\pm$0.39 & 90.73$\pm$0.50 & 90.63$\pm$0.57 & 98.21$\pm$0.18 & 0.229$\pm$0.034 & 87.58$\pm$0.79\\
\base+VC & 93.09$\pm$0.26 & 91.35$\pm$0.43 & 91.33$\pm$0.39 & 98.41$\pm$0.34 & 0.220$\pm$0.060 & 89.94$\pm$0.26\\
\method (Ours) & \textbf{93.33$\pm$0.24} & \textbf{91.71$\pm$0.27} & \textbf{91.69$\pm$0.29} & \textbf{98.80$\pm$0.07} & \textbf{0.153$\pm$0.013} & \textbf{90.97$\pm$0.37}\\
\bottomrule
\end{tabular}}
\end{table}

\subsection{Robustness Diagnostics}
\datasetc\ evaluates seven feature-level corruption families: pose dropout, coordinate jitter, temporal downsampling, low confidence, feature quantization, temporal feature smoothing, and upper-limb feature suppression. Each corruption has three severity levels and preserves the original label. The protocol is applied to processed pose/RGB features so that every method receives the same degraded input, which isolates recognition robustness from stochastic feature-extraction variability and makes the benchmark reproducible.
\begin{table}[t]
\centering
\setlength{\belowcaptionskip}{6pt}
\vspace*{8pt}
\vspace*{1pt}
\caption{\datasetc\ corruption families. Severity increases the amount of perturbation while preserving the original label.}
\label{tab:corruptions}
\resizebox{\linewidth}{!}{
\begin{tabular}{lll}
\toprule
Corruption & Severity values & Degraded signal\\
\midrule
Pose dropout & $p=\{0.08,0.18,0.30\}$ & Random low-confidence joints with coordinate noise\\
Coordinate jitter & $\sigma=\{0.035,0.075,0.13\}$ & Perturbed normalized joint coordinates\\
Temporal downsampling & step $\{2,3,4\}$ & Repeated/skipped temporal evidence\\
Low confidence & scale $\{0.78,0.55,0.36\}$ & Reduced confidence with small coordinate noise\\
Feature quantization & grid $\{96,48,24\}$ & Quantized pose coordinates and reduced confidence\\
Temporal feature smoothing & $\alpha=\{0.25,0.45,0.65\}$ & Smoothed pose/RGB dynamics and velocity attenuation\\
Upper-limb feature suppression & conf. scale $\{0.60,0.32,0.12\}$ & Suppressed shoulder, elbow, and wrist evidence\\
\bottomrule
\end{tabular}}
\end{table}

\begin{figure}[t]
\centering
\includegraphics[width=.43\linewidth]{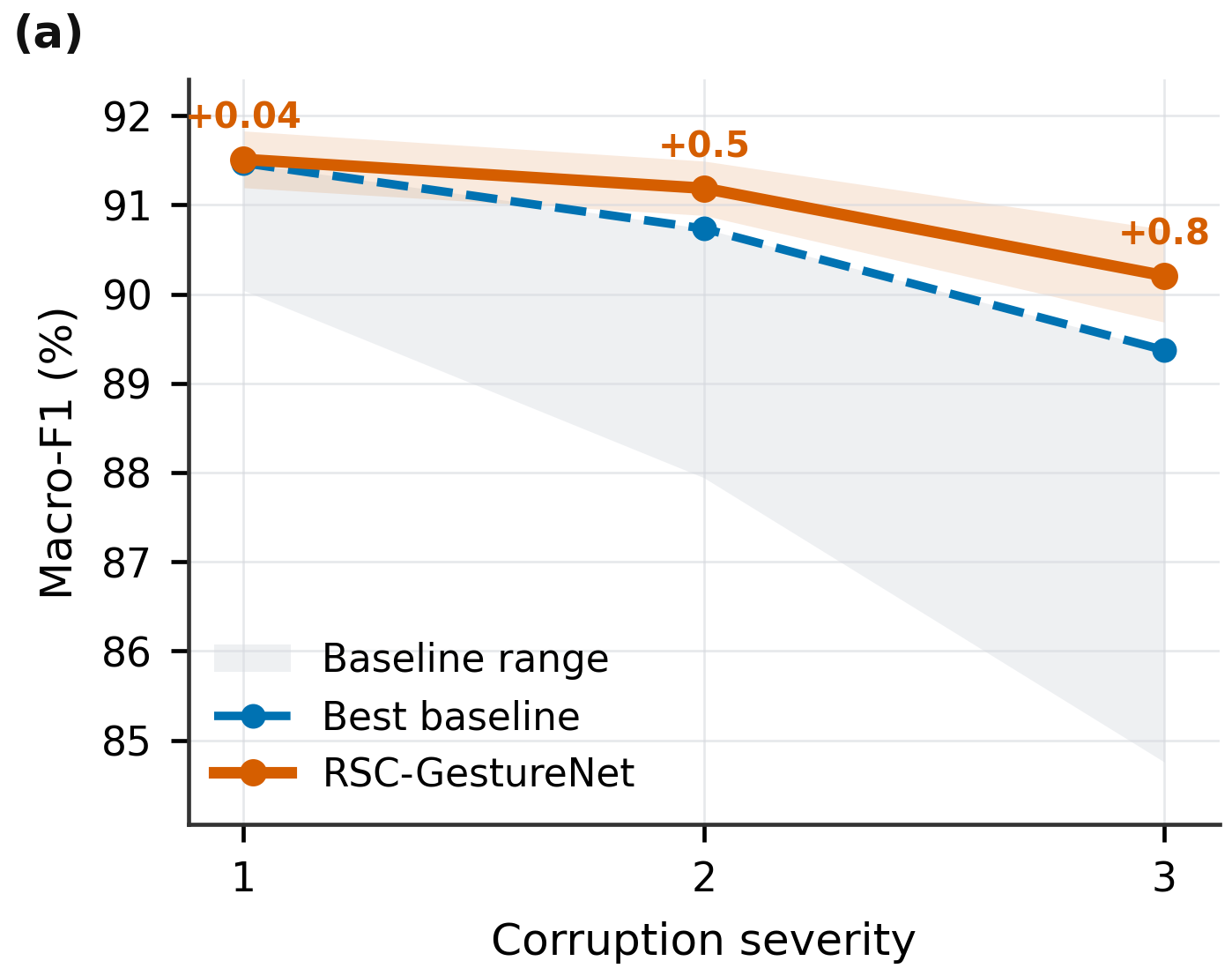}
\includegraphics[width=.43\linewidth]{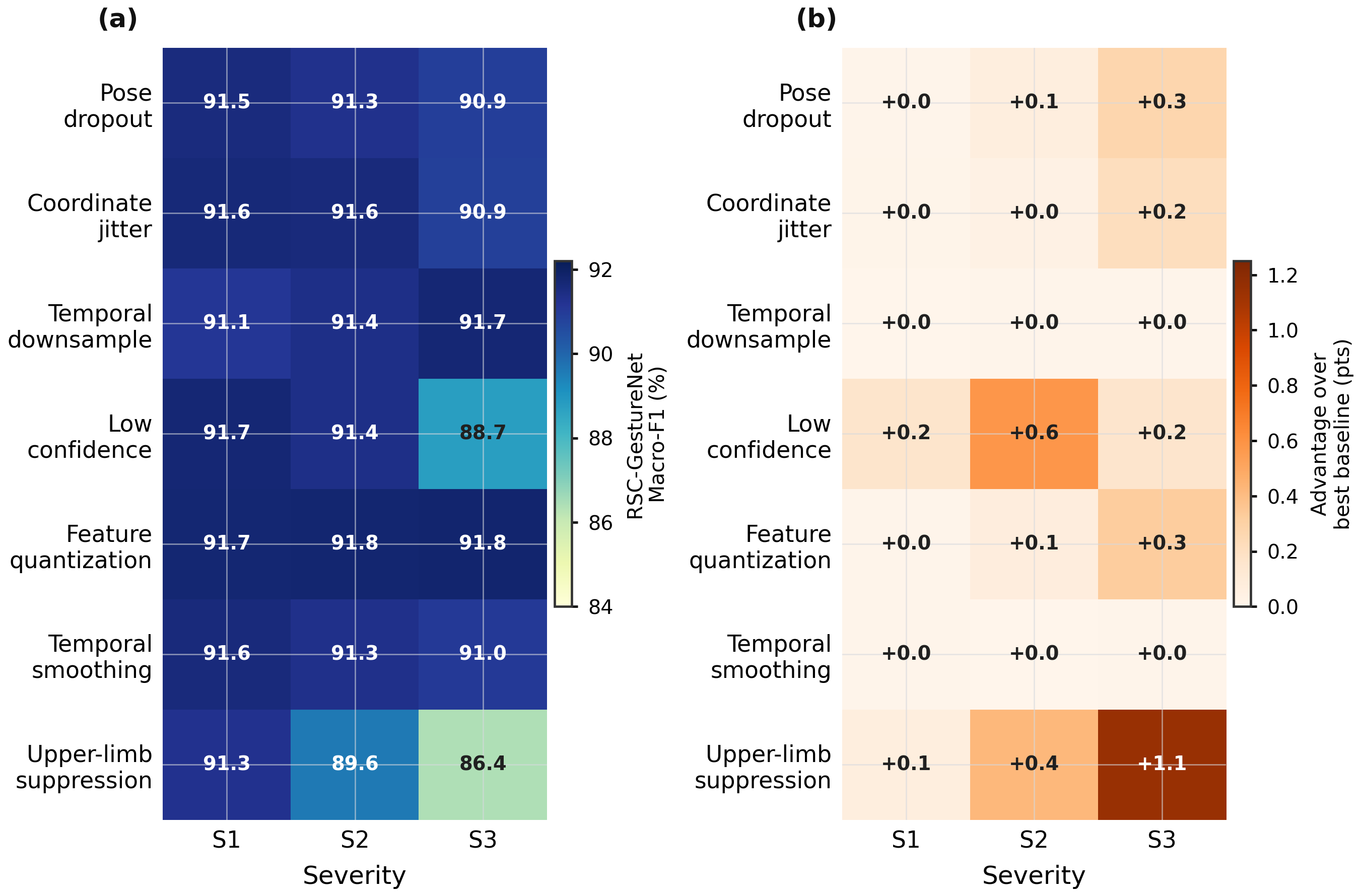}
\caption{Robustness analysis on \datasetc. Left: macro-F1 by corruption severity. Right: per-corruption macro-F1 and its advantage over the strongest corresponding baseline.}
\label{fig:robust}
\end{figure}

Figure~\ref{fig:robust} shows that \method\ obtains the highest mean robust macro-F1 at each severity among robustness-profiled methods, reaching 91.51\%, 91.19\%, and 90.21\% at severities 1--3. The per-corruption heatmap shows gains over the strongest corresponding baseline across all 21 corruption-severity settings, with the largest gains under upper-limb suppression and pose dropout. This behavior is consistent with the learned reliability trigger: as shown in Table~\ref{tab:statsbranch}, the fusion branch dominates on clean data, but upper-limb suppression shifts probability mass toward the TCN and pose-only fallbacks, preventing a fixed fusion branch from over-trusting corrupted arm evidence.

Because feature-level perturbations are deterministic and reproducible but do not re-run pose extraction, we add a separate RGB-level diagnostic. Original RGB frames are corrupted, MediaPipe pose is re-extracted, and recognition is evaluated on a representative subset of three test videos with 600 frames each and three severity levels. This subset is not merged with the full-test Robust-F1 score. As shown in Table~\ref{tab:rgbdiag}, \arvariant\ is strongest under low light and obtains the best macro-F1 under motion blur, while remaining within 0.27 online-F1 points of the best online model under blur. Thus the image-level experiment supports a conservative claim: adaptive reliability improves common illumination/blur stress cases after pose re-extraction, rather than establishing a comprehensive image-corruption SOTA.

\begin{table}[t]
\centering
\setlength{\belowcaptionskip}{6pt}
\vspace*{5pt}
\vspace*{1pt}
\caption{RGB/image-level corruption diagnostic with MediaPipe pose re-extraction. Values are macro-F1/online-F1 on the diagnostic subset; the clean full-test comparison remains Table~\ref{tab:main}.}
\label{tab:rgbdiag}
\scriptsize
\setlength{\tabcolsep}{4pt}
\begin{tabular}{lcccc}
\toprule
Image corruption & Best Macro-F1 & Best Online-F1 & Ours Macro-F1 & Ours Online-F1\\
\midrule
Low light & \textbf{80.98} (ours) & \textbf{80.92} (ours) & \textbf{80.98} & \textbf{80.92}\\
Motion blur & \textbf{87.97} (ours) & 88.32 (ST-GCN) & \textbf{87.97} & 88.05\\
\bottomrule
\end{tabular}
\end{table}

\subsection{Ablation Study}
Table~\ref{tab:ablation} isolates the pre-calibrated RSC-GestureNet-Fusion backbone. Removing pose--RGB fusion reduces macro-F1 from 90.28\% to 88.49\% and nearly doubles TTC, showing that appearance cues complement pose. Removing reliability gating or the dynamic graph causes similar degradation, while removing the danger-aware loss produces the largest drop. Removing stabilization emits earlier but increases PSR to 1.60/s, confirming that early response should be balanced against command flicker.

\begin{table}[H]
\centering
\setlength{\belowcaptionskip}{6pt}
\vspace*{5pt}
\vspace*{1pt}
\caption{Ablation study on the official CTPGesture v1 split under the same causal protocol. $\metricup$ indicates higher is better and $\metricdown$ indicates lower is better.}
\label{tab:ablation}
\scriptsize
\setlength{\tabcolsep}{4pt}
\renewcommand{\arraystretch}{0.95}
\resizebox{0.96\linewidth}{!}{
\begin{tabular}{lccccc}
\toprule
Variant & Accuracy (\%)$\metricup$ & Macro-F1 (\%)$\metricup$ & Early@10 (\%)$\metricup$ & TTC (s)$\metricdown$ & PSR (/s)$\metricdown$\\
\midrule
Full RSC-GestureNet-Fusion & 91.71 & 90.28 & 98.84 & 0.172 & 0.00\\
No stabilization & 91.71 & 90.28 & 99.19 & 0.086 & 1.60\\
Pose only, no RGB & 89.86 & 88.49 & 98.25 & 0.303 & 0.00\\
No reliability gating & 89.83 & 88.26 & 98.25 & 0.250 & 0.00\\
No dynamic graph & 89.72 & 88.20 & 98.49 & 0.217 & 0.00\\
No boundary term & 88.79 & 87.32 & 98.37 & 0.247 & 0.00\\
No danger-aware loss & 88.36 & 86.92 & 98.14 & 0.336 & 0.00\\
\bottomrule
\end{tabular}}

\end{table}

\subsection{Online Behavior and Qualitative Analysis}
\begin{figure}[t]
\centering
\includegraphics[width=.43\linewidth]{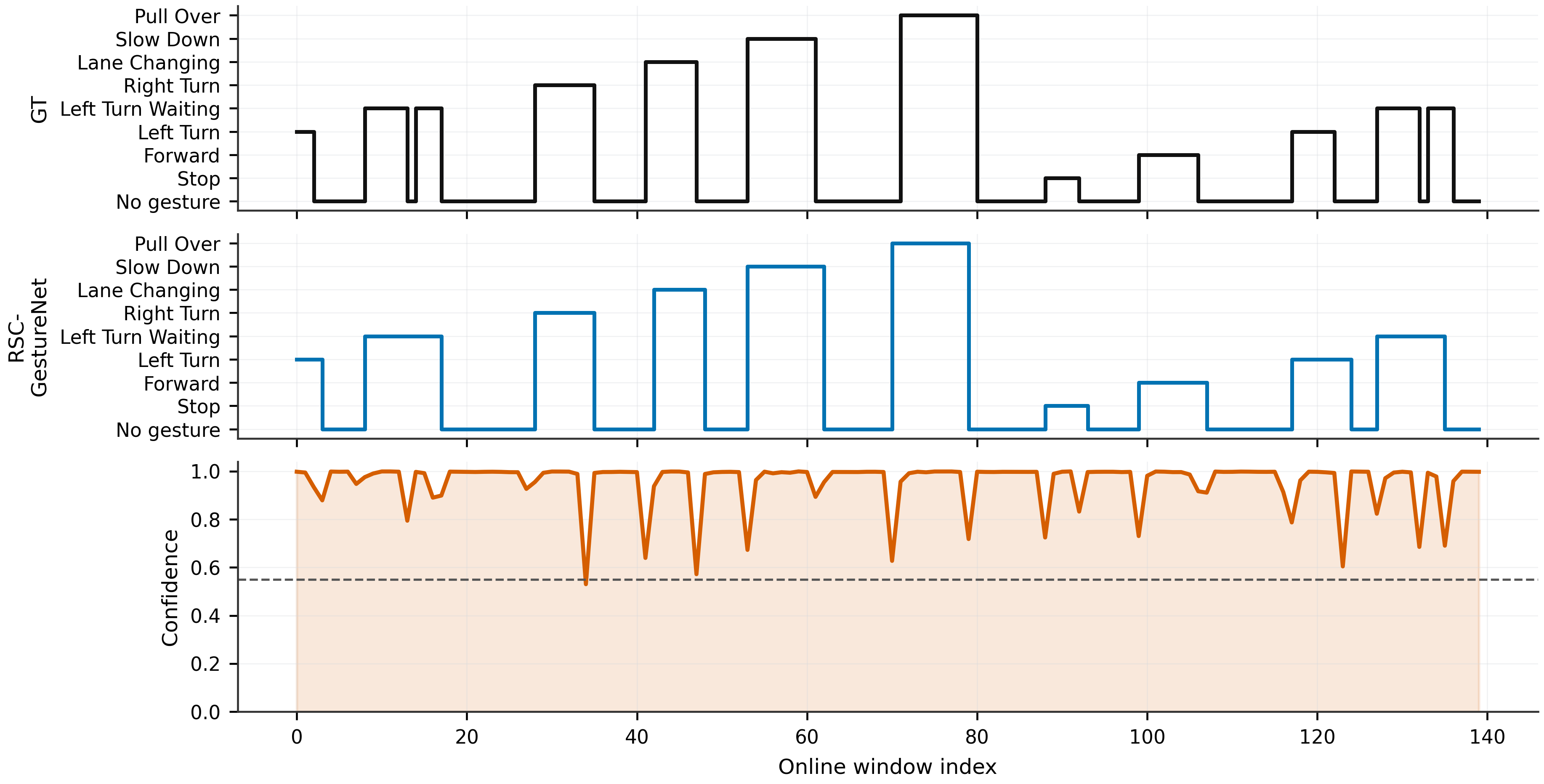}
\includegraphics[width=.43\linewidth]{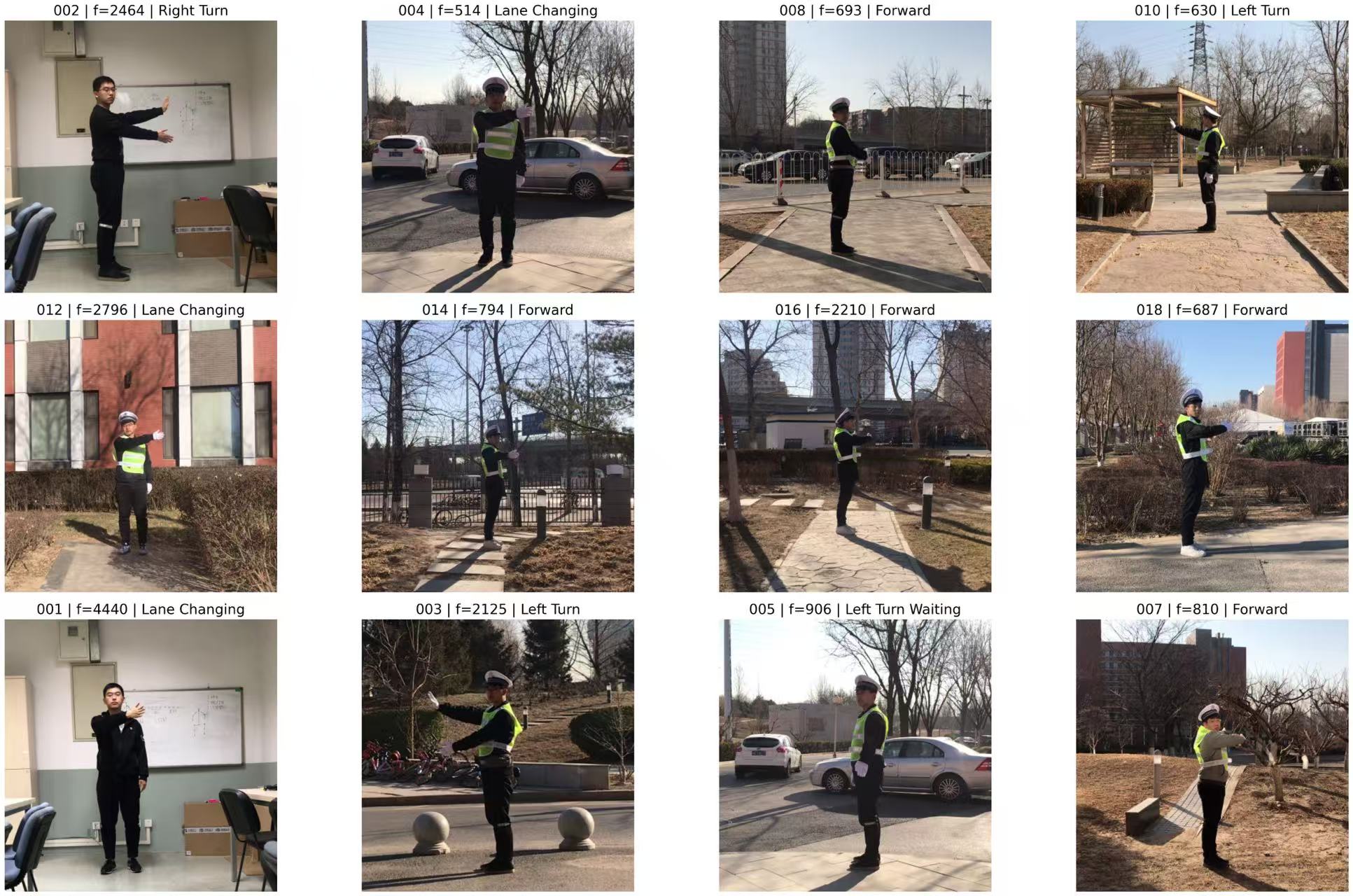}
\caption{Qualitative evaluation. Left: online prediction timeline on a continuous test segment. Right: representative full-frame samples from the official CTPGesture v1 videos.}
\label{fig:qual}
\end{figure}

Online evaluation measures both frame-level correctness and command stability. Figure~\ref{fig:qual} visualizes a continuous test segment: compared with TCN and ST-GCN, \method\ tracks transitions with fewer delayed corrections and a steadier command stream. The raw frames show that the officer occupies only part of a road scene, so discriminative evidence must be integrated across body pose, arm motion, and coarse appearance.

\subsection{Selective Output, Calibration, and Error Structure}
\begin{figure}[H]
\centering
\begin{minipage}[t]{.70\linewidth}
\vspace{0pt}
\centering
\includegraphics[width=\linewidth]{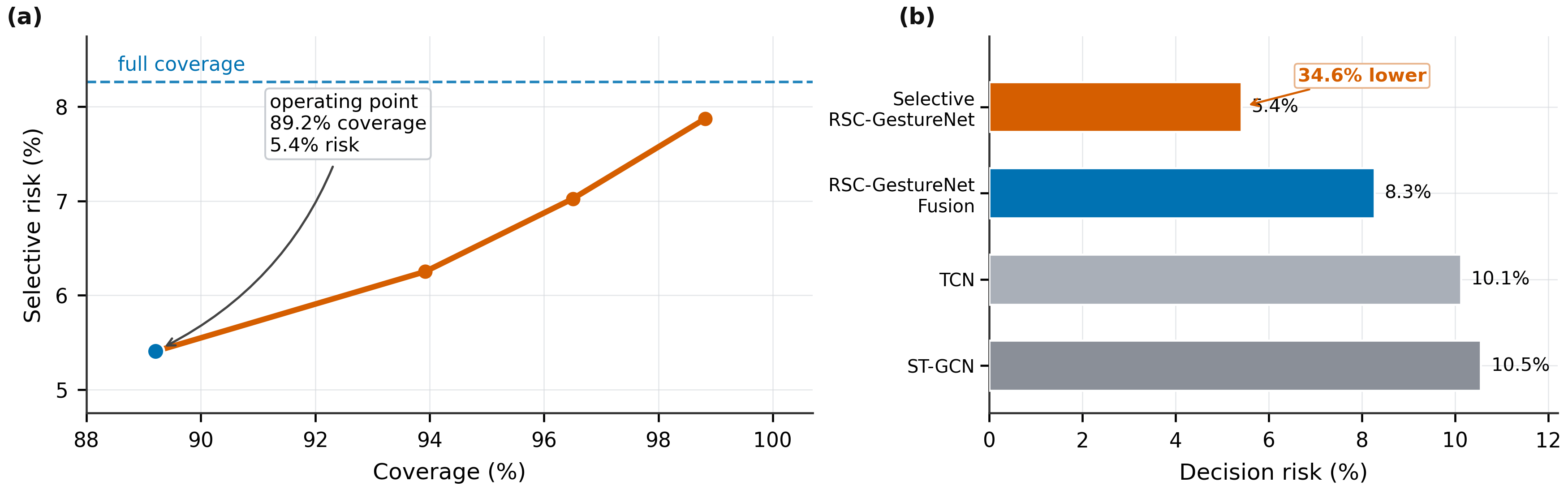}
\end{minipage}\hfill
\begin{minipage}[t]{.28\linewidth}
\vspace*{3pt}
\centering
\includegraphics[width=\linewidth]{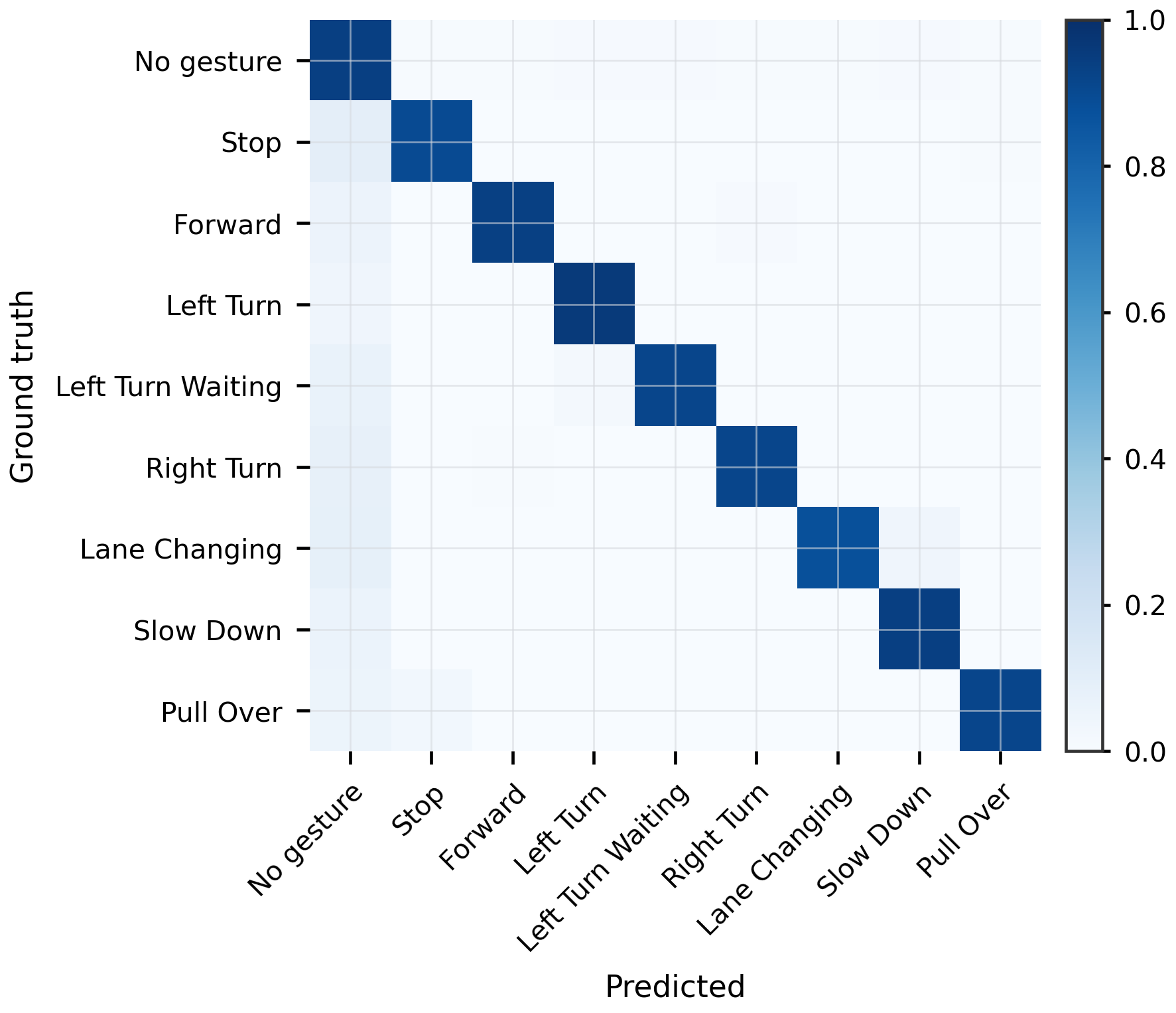}
\end{minipage}
\par\vspace*{4pt}
\makeatletter
\long\def\@makecaption#1#2{%
  \small
  \vskip\abovecaptionskip
  \noindent{\bfseries #1.} #2\par
  \vskip\belowcaptionskip}
\makeatother
\caption{(Left) Risk--coverage diagnostics of \method. Panel (a) shows selective risk as coverage varies and marks the selected operating point; panel (b) compares decision risk across the evaluated models.}
\label{fig:riskcoverage}
\vspace*{-2pt}
\caption{(Right) Class-level confusion matrix of \method\ on the official test split.}
\label{fig:confusion}
\end{figure}

Selective output provides a safety-oriented interface above the classifier: high-confidence, temporally consistent predictions are emitted, whereas unstable windows are deferred. Figure~\ref{fig:riskcoverage} shows the expected risk--coverage curve for \method: increasing conservatism lowers selective risk at the cost of coverage. To make the comparison fair, Table~\ref{tab:calibselect} evaluates all calibrated methods at the same coverage and the same risk levels. \method\ has the lowest selective risk at 90\%, 95\%, and 99\% coverage and the highest coverage under 6\% and 7\% risk; at 8\% risk all compared methods reach 100\% coverage. The same table also shows that \method\ has the lowest Brier score and NLL, indicating better whole-distribution probability quality rather than merely a more conservative abstention rule.

Figure~\ref{fig:confusion} shows that residual errors mainly occur among visually similar turning and waiting commands. For Stop, the dominant residual output is inactive rather than a motion command (Table~\ref{tab:perclass}), which is consistent with the danger-aware loss and selective emission near boundaries.
\FloatBarrier

\begin{table}[H]
\centering
\setlength{\belowcaptionskip}{4pt}
\vspace*{1pt}
\caption{Calibration quality and fair selective-output comparison. Risk@Cov reports selective risk at fixed coverage; Cov@Risk reports coverage at fixed selective risk. Values are mean $\pm$ standard deviation over three seeds. Best values are bold.}
\label{tab:calibselect}
\scriptsize
\setlength{\tabcolsep}{2.7pt}
\resizebox{\linewidth}{!}{
\begin{tabular}{lccccccc}
\toprule
Model & Brier$\downarrow$ & NLL$\downarrow$ & Risk@90$\downarrow$ & Risk@95$\downarrow$ & Risk@99$\downarrow$ & Cov@6$\uparrow$ & Cov@7$\uparrow$\\
\midrule
\method & \textbf{0.1007$\pm$0.0037} & \textbf{0.2065$\pm$0.0066} & \textbf{3.16$\pm$0.27} & \textbf{4.56$\pm$0.22} & \textbf{6.23$\pm$0.26} & \textbf{98.49$\pm$0.55} & \textbf{100.00$\pm$0.00}\\
\base+VC & 0.1044$\pm$0.0033 & 0.2073$\pm$0.0119 & 3.49$\pm$0.17 & 4.97$\pm$0.16 & 6.46$\pm$0.22 & 97.74$\pm$0.62 & 99.87$\pm$0.23\\
TCN+VC & 0.1078$\pm$0.0014 & 0.2160$\pm$0.0035 & 3.66$\pm$0.01 & 5.12$\pm$0.08 & 6.61$\pm$0.18 & 97.34$\pm$0.45 & 99.79$\pm$0.36\\
ST-GCN+VC & 0.1140$\pm$0.0058 & 0.2219$\pm$0.0124 & 4.08$\pm$0.40 & 5.52$\pm$0.43 & 7.09$\pm$0.37 & 96.11$\pm$1.20 & 98.68$\pm$0.96\\
\bottomrule
\end{tabular}}
\vspace*{1pt}
\end{table}

\FloatBarrier
\subsection{Statistical Reliability and Adaptive Branching}
The calibration and selective-output procedure is designed to avoid test-set leakage. Vector calibration, online threshold selection, and branch-weight selection use only the training-internal stream; the official test split is used once for final reporting. Table~\ref{tab:statsbranch} summarizes two labeled diagnostics. Panel (a) reports matched three-seed tests, showing statistically significant gains in accuracy, online-F1, Early@10, and TTC, while macro-F1 remains a stable positive trend. Panel (b) verifies the intended reliability trigger: when severe upper-limb evidence is suppressed, the model automatically reduces the fusion weight and increases fallback contributions.

\begin{table}[H]
\centering
\setlength{\belowcaptionskip}{6pt}
\vspace*{5pt}
\vspace*{1pt}
\caption{Statistical significance and adaptive branch-weight diagnostics for \method. Panel (a) evaluates matched three-seed gains; panel (b) reports reliability-triggered branch weights.}
\label{tab:statsbranch}
\scriptsize
\setlength{\tabcolsep}{2.5pt}
\begin{minipage}{.52\linewidth}
\centering
\textbf{(a) Three-seed matched tests}\\[-1pt]
\resizebox{\linewidth}{!}{%
\begin{tabular}{lccc}
\toprule
Metric & Comparison & Gain & $p$\\
\midrule
Accuracy & vs. \base+VC & +0.24 pts & 0.005\\
Macro-F1 & vs. \base+VC & +0.35 pts & 0.066\\
Online-F1 & vs. \base+VC & +0.36 pts & 0.026\\
Early@10 & vs. TCN+VC & +0.39 pts & 0.038\\
TTC & vs. TCN+VC & 0.061 s faster & 0.027\\
\bottomrule
\end{tabular}}
\end{minipage}\hfill
\begin{minipage}{.46\linewidth}
\centering
\textbf{(b) Adaptive branch weights}\\[-1pt]
\resizebox{\linewidth}{!}{%
\begin{tabular}{lccc}
\toprule
Condition & Fusion & TCN & Pose\\
\midrule
Clean & 0.992$\pm$0.043 & 0.002$\pm$0.022 & 0.005$\pm$0.022\\
UL-S3 & 0.608$\pm$0.053 & 0.225$\pm$0.052 & 0.167$\pm$0.022\\
\bottomrule
\end{tabular}}
\end{minipage}
\end{table}

Table~\ref{tab:perclass} reports per-class precision, recall, F1, and the dominant residual prediction for \method. Most command classes exceed 90\% F1. The lowest F1 occurs for Stop because the model occasionally maps transition regions to the inactive/no-command class; importantly, the dominant residual error for Stop is inactive output rather than a motion command. Lane Changing and Left Turn Waiting remain difficult because their arm configurations often occur near transition boundaries or under partial upper-limb occlusion.

\begin{table}[H]
\centering
\setlength{\abovecaptionskip}{9pt}
\setlength{\belowcaptionskip}{6pt}
\vspace*{1pt}
\caption{Per-class error analysis and dominant residual predictions for \method\ on the official CTPGesture v1 test split. Precision, recall, and F1 are mean $\pm$ standard deviation over three seeds; the inactive class is the official ``No gesture'' label and is distinct from selective abstention.}
\label{tab:perclass}
\scriptsize
\setlength{\tabcolsep}{3pt}
\resizebox{\linewidth}{!}{
\begin{tabular}{lccccc}
\toprule
Class & Support & Precision (\%)$\metricup$ & Recall (\%)$\metricup$ & F1 (\%)$\metricup$ & Main residual prediction\\
\midrule
Inactive & 9484 & 95.31$\pm$0.33 & 93.93$\pm$0.09 & 94.62$\pm$0.21 & Left Turn Waiting (0.93\%)\\
Stop & 513 & 81.03$\pm$0.96 & 90.97$\pm$1.14 & 85.71$\pm$0.11 & Inactive (8.51\%)\\
Forward & 983 & 93.46$\pm$0.20 & 94.07$\pm$0.92 & 93.76$\pm$0.55 & Inactive (5.39\%)\\
Left Turn & 911 & 89.97$\pm$0.96 & 96.08$\pm$0.17 & 92.92$\pm$0.59 & Inactive (3.92\%)\\
Left Turn Waiting & 963 & 90.83$\pm$0.80 & 90.76$\pm$0.63 & 90.79$\pm$0.15 & Inactive (7.27\%)\\
Right Turn & 910 & 91.51$\pm$0.18 & 93.59$\pm$2.09 & 92.53$\pm$1.01 & Inactive (5.64\%)\\
Lane Changing & 858 & 93.47$\pm$0.28 & 88.42$\pm$0.71 & 90.88$\pm$0.33 & Inactive (7.81\%)\\
Slow Down & 1079 & 89.94$\pm$0.92 & 94.01$\pm$0.99 & 91.92$\pm$0.18 & Inactive (5.99\%)\\
Pull Over & 1057 & 92.96$\pm$0.27 & 91.55$\pm$0.05 & 92.25$\pm$0.11 & Inactive (4.98\%)\\
\bottomrule
\end{tabular}}
\end{table}
\FloatBarrier

\section{Discussion}
The direct traffic-graph baselines and RGB-level pose re-extraction diagnostic address two validity concerns: domain-specific comparison and robustness beyond feature perturbation. The results support the central hypothesis that traffic-police gesture recognition benefits from modeling pose reliability instead of treating all detected joints as equally trustworthy. Calibration, selective-output, significance, and branch-weight checks show that the gains are not merely due to abstention or seed noise. Remaining errors concentrate around visually similar turning/waiting gestures near transition boundaries and uncertain upper-limb evidence, suggesting that future work should combine reliability-aware recognition with richer phase modeling and planner-level uncertainty handling.

\vspace{2pt}

\section{Conclusion}
This study presented \method\ for reliability-aware selective causal recognition of Chinese traffic police gestures. On the complete official CTPGesture v1 split under a unified full-frame causal protocol, it achieves the best mean performance among the evaluated methods across clean, online, early-decision, command-delay, calibration, selective-risk, and robustness metrics. Direct MD-GCN/HLP-GCN reimplementations and RGB-level pose re-extraction diagnostics further support the deployment value of reliability-aware modeling. Beyond improving recognition scores, the study provides a reproducible evaluation protocol and a calibrated deferrable command interface that can help bridge traffic-police gesture perception with safety-oriented autonomous-driving planning.
\renewcommand{\doi}[1]{}
\bibliographystyle{splncs04}
\bibliography{references}

\end{document}